\documentclass[conference]{IEEEtran}
\IEEEoverridecommandlockouts
\usepackage{cite}
\usepackage{amsmath,amssymb,amsfonts}
\usepackage{algorithmic}
\usepackage{textcomp}
\usepackage{xcolor}
\usepackage{todonotes}
\usepackage{lipsum}
\usepackage{float}
\def\BibTeX{{\rm B\kern-.05em{\sc i\kern-.025em b}\kern-.08em
    T\kern-.1667em\lower.7ex\hbox{E}\kern-.125emX}}

\usepackage{graphicx}
\graphicspath{{./images/}}

\usepackage{booktabs} 
\usepackage{array}

\usepackage[english]{babel}

\begin{document}

\title{Contextual Categorization Enhancement\\through LLMs Latent-Space}

\author{\IEEEauthorblockN{Zineddine Bettouche, Anas Safi, Andreas Fischer}
    \IEEEauthorblockA{Deggendorf Institute of Technology \\
        Dieter-Görlitz-Platz 1, 94469 Deggendorf \\
        zineddine.bettouche@th-deg.de, anas.safi@stud.th-deg.de, andreas.fischer@th-deg.de}
}

\maketitle

\begin{abstract}
    Managing the semantic quality of the categorization in large textual datasets, such as Wikipedia, presents significant challenges in terms of complexity and cost. In this paper, we propose leveraging transformer models to distill semantic information from texts in the Wikipedia dataset and its associated categories into a latent space. We then explore different approaches based on these encodings to assess and enhance the semantic identity of the categories. Our graphical approach is powered by Convex Hull, while we utilize Hierarchical Navigable Small Worlds (HNSWs) for the hierarchical approach. As a solution to the information loss caused by the dimensionality reduction, we modulate the following mathematical solution: an exponential decay function driven by the Euclidean distances between the high-dimensional encodings of the textual categories. This function represents a filter built around a contextual category and retrieves items with a certain Reconsideration Probability (RP). Retrieving high-RP items serves as a tool for database administrators to improve data groupings by providing recommendations and identifying outliers within a contextual framework.
\end{abstract}

\begin{IEEEkeywords}
    Natural Language Processing, Contextual Categorization, Large Language Models, BERT, Convex Hull, Hierarchical Navigable Small Worlds, High-dimensional Latent Space, Dimensionality Reduction.
\end{IEEEkeywords}

\section{Introduction}

In the modern age of abundant textual data, effective categorization poses a significant challenge due to its increasing complexity. As data volumes grow exponentially, traditional methods struggle to handle the task adequately.

Fortunately, advancements in Natural Language Processing (NLP) provide a promising solution. NLP techniques, like word embeddings, encode semantic information, enabling automated and accurate categorization of vast datasets. Long Short-Term Memory (LSTM) networks, a type of recurrent neural network, excel at modeling sequential data and have proven effective in automating categorization tasks.

Transformer models~\cite{transformers}, such as the BERT series~\cite{bert}, offer a deep understanding of contextual nuances, enhancing categorization accuracy. Innovative algorithms like Convex Hull~\cite{Graham} and Hierarchical Navigable Small World (HNSW)~\cite{malkov}, powered by embeddings, can be employed to check the categorization efficiency by organizing and navigating through high-dimensional spaces.

In our previous works~\cite{bettouche2022conference, bettouche2023topical}, we addressed the contextual clustering of transformer encodings in an unlabeled database of scientific articles. In this study, we investigate the effectiveness of these methodologies—NLP techniques, LLMs, and geometric algorithms—in improving categorization efficiency and accuracy. Through empirical analysis, we aim to demonstrate their efficacy in managing large-scale data categorization challenges.

As for the structure of this paper, Section II offers a background on key topics such as Transformer models, and the NLP techniques employed. Section III cites the related works, setting a foundation and context for the research. Section IV lays out the methodology, detailing the approach, models, and metrics used. Section V presents the experiments, the challenges faced, trade-offs considered, and the results derived from the techniques employed. Finally Section VI concludes the study and sets up future work.

\section{Background}\label{background}
This section introduces the background of the techniques used to develop our methodology.

\subsection{Transformer Models: BERT}
Transformer models, such as BERT, are pivotal tools in the field of natural language processing. BERT, which stands for Bidirectional Encoder Representations from Transformers, has transformed the way we handle text data. BERT models are pretrained on vast textual corpora and excel at converting text into high-dimensional vectors. They shine at capturing intricate semantic relationships between words and sentences, making them invaluable for various NLP tasks.

\subsection{Convex Hull}
Convex hulls are key concepts in computational geometry and mathematics. They represent a closed, convex shape that encloses a set of data points in multi-dimensional space. In our study, we use convex hulls to define boundaries around groups of BERT-encoded articles, creating distinct regions within the latent space. This method aids in exploring how articles are distributed within a specific category and identifying those that are positioned near the boundaries. This enables us to adjust and extend category definitions effectively.

\subsection{Hierarchical Navigable Small Worlds}
Hierarchical Navigable Small Worlds (HNSWs) serve as a data structure for approximate similarity searches in high-dimensional spaces. They offer a practical and scalable means to navigate complex, multi-dimensional data while preserving proximity relationships. In our research, HNSWs are employed to efficiently organize and search BERT-encoded vectors. This facilitates the process of identifying articles that closely resemble a given category within the latent space. The hierarchical nature of HNSWs enhances our ability to adjust and expand categories based on latent similarities.

\section{Related Work}\label{relatedWork}
In this section, we provide a concise overview of related work, highlighting key contributions that inform our study. 

Moas' investigation into ``Real-time prediction of Wikipedia articles’ quality"~\cite{maos} examines various assessment methods, automated and manual, highlighting the challenges in manual assessment due to its time-consuming nature and issues like ``circular categories." The paper suggests using an Application Programming Interface (API) to address these challenges in analyzing category trees. Another study, ``The Use of NLP-Based Text Representation Techniques to Support Requirement Engineering Tasks,"~\cite{sonbol2022nlp} discusses levels of Natural Language Processing (NLP). It emphasizes the semantic level for understanding text meaning and supports the adoption of the Vector Space Model~\cite{vsm} for its simplicity in representing articles in high-dimensional spaces.

The paper ``Data Mining with Python"~\cite{nilsen} by Nielsen highlights NetworkX and DiGraphs' utility in analyzing hierarchical structures like category trees in Wikipedia. These directed graphs accurately represent parent-child relationships, essential in modeling category relationships. Studies on convex hulls, such as Preparata and Hong's computational aspects~\cite{Preparata} and Graham's algorithm~\cite{Graham}, provide insights into computational methods for points in two and three-dimensional spaces. Understanding convex hull derivation from simple polygons aids in comprehending data distribution in high-dimensional spaces, relevant to machine learning. 

The research by Malkov and Yashunin~\cite{malkov} introduces Hierarchical Navigable Small World (HNSW) graphs for kNN search, implemented in this thesis for vector retrieval. The analysis revealed that articles within the same category tend to be the closest neighbors in the constructed HNSW structure. 

Johnson et al.~\cite{Johnson} proposed a language-agnostic method for categorizing Wikipedia articles, overcoming content quality and geographic variations. Leveraging article links, their approach matches language-dependent methods in performance, extending coverage across Wikipedia languages. Biswas et al.~\cite{Biswas} introduced Cat2Type, enhancing entity typing in knowledge graphs (KGs) like DBpedia and Freebase using Wikipedia categories. By utilizing semantic information, Cat2Type surpasses existing methods on real-world datasets. Ostendorff et al.~\cite{Ostendorff} addressed semantic relationship identification between documents, treating it as a pairwise classification task. Employing BERT and XLNet, experiments on Wikipedia article pairs and Wikidata properties validated BERT's effectiveness, suggesting potential for semantic document-based recommender systems.

This overview summarizes findings from various research domains, including article categorization, NLP techniques, dimensionality reduction, graph analysis, convex hulls, and nearest neighbor search, all relevant to this study. Notably, the reviewed research does not directly address the use of transformer-generated encodings in their high-dimensionality (full information), which differentiates the methodology presented in this paper. The paper introduces a novel mathematical function that utilizes these encodings, representing a distinctive contribution to the field.

\section{Methodology}\label{methodology}

This section presents the Wikipedia dataset used in this work and the approaches developed to select articles for context-based reconsideration.

\subsection{Overview of Wikipedia Dumps}
\begin{figure}
    \centering
    \includegraphics[width=.97\linewidth]{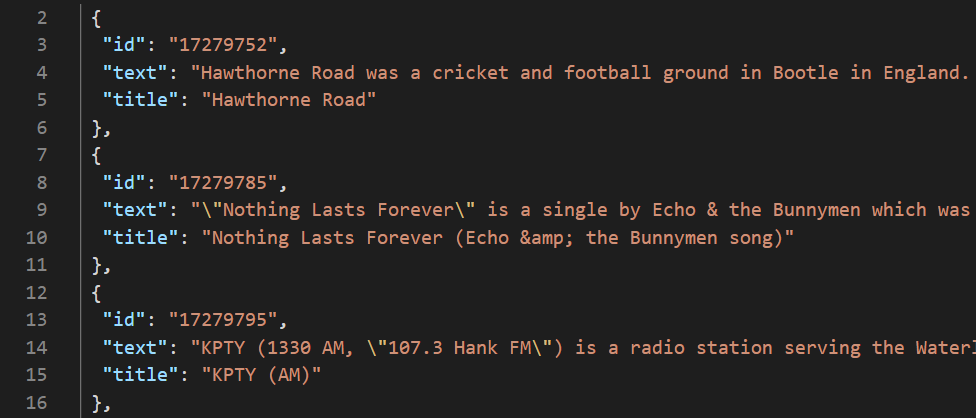}
	\caption{Sample of Data Objects in the Wikipedia Dataset}    
	\label{fig:data-object}
\end{figure}

The data used is from the Wikipedia dump of November 2020 on Kaggle~\cite{dump} (plain text). There are 6,144,363 articles in the dataset divided into 605 JSON files. Figure~\ref{fig:data-object} shows a random sample of items in these JSON files. Each item has an id, a text, and a title.

The transformer model processes articles individually, ensuring consistent and undistorted encodings regardless of the number of articles processed and sample size. Therefore, as a proof of concept and for computational feasibility, we randomly selected 10,000 articles to assess the approaches and conduct experiments.

\subsection{Convex Hull}
Our second approach involves the construction of a convex hull on the set of vectors representing the category. The vectors of the category tree are mapped onto a 2D plane using UMAP~\cite{UMAP}. The reason behind including UMAP is that it is not feasible to construct a convex hull in the 768 dimensions Every article in the dataset undergoes encoding and mapping onto the 2D plane, enabling the determination of whether any article breaches the established convex hull boundary. Convex hulls provide an efficient way to capture the spatial relationships between category vectors when mapped onto a 2D plane. This approach leverages the geometric properties of convex hulls, enabling the visualization and definition of the boundaries of category clusters. By determining whether an article breaches the established convex hull boundary, one can promptly identify articles that do not conform to their designated categories.

\subsection{HNSW}
The application of Hierarchical Navigable Small World (HNSW) is used in the third approach to retrieve the nearest neighbors of articles within a category tree. The premise of this approach is that retrieved articles should already share the same category. The presence of articles that are retrieved but do not belong to the category tree should be reconsidered for category addition. We construct an HNSW using the randomly selected articles, alongside category vectors. For each category vector, we retrieve the five nearest neighbors in this HNSW. Articles that appear and are not associated with the category, are flagged for inspection. This approach leverages the structure of the data. Articles within the same category should naturally be closer to each other in the vector space. By flagging articles retrieved by HNSW that do not belong to the category tree, one can promptly identify instances of lack of categorization.

\subsection{Filter built on High-dimensional Latent Space}
\label{method:filter}
The high dimensionality of the encodings can be too complex for several techniques, and it is referred to as \textit{the curse of dimensionality} in this case. However, we intend to use these information-rich vectors to our advantage. We design a filter that takes the category articles (as encodings) and its centroid vector. This filter is applied on the articles of the sample. It takes a percentage that represents the Reconsideration Probability (RP). 

An article with 100\% RP must be reconsidered to be added to the concerned category. We encode the category into latent space and calculate the centroid vector. The category encodings form a cloud around this centroid vector in the latent space. Any non-category vector in the dataset that has a distance to the centroid vector equal to or less than the radius of this category cloud is assigned 100\% RP.

We assume that in the sample there must at least one article that must not be in the selected category. We assign this non-category article 0.1\% RP (0\% RP complicates the mathematical modeling of the filter). In the case that the dataset admin cannot select such an article, we assign 0.1\% RP to the article with the farthest encoding from the centroid. 

As for the filter function, let $d_{c}$ represent the distance of the last quarter threshold (75th percentile). This value is often used as a measure of central tendency that is less affected by outliers compared to the mean. Let $d_{ea}$ represent the distance of the examined article. The $RP(d_{c}, d_{ea})$ should be 100\% when $d_{ea}$ is equal to $d_{c}$, and approach 0\% the greater is $d_{ea}$ compared to $d_{c}$. We achieve this is by using an exponential decay function with a horizontal shift. To determine the decay constant $k$, we need to consider the set of non-category articles and their distances. The median value $median(set(d_{ea}))$ in this set of distances should result in 50\% RP. Equations~\ref{eq:k} and \ref{eq:rp} show the mathematical modeling of this description.

\begin{equation}
    k = \frac{-ln(0.5)}{median(set(d_{ea}))-d_{c}}
    \label{eq:k}
\end{equation}

\begin{equation}
    RP(d_{c}, d_{ea}) = 100*e^{-k*(d_{ea}-d_{c})} 
    \label{eq:rp}
\end{equation}

\section{Experiments}\label{experiments}
This section discusses the experiments done in this paper.

\subsection{Setting up the Vector Space: Encodings of the Category and the Sample Articles}
\begin{figure}
    \centering
    \includegraphics[width=.97\linewidth]{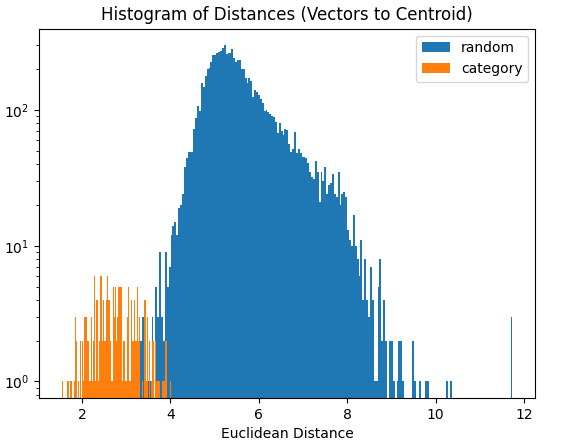}
    \caption{Euclidean Distances between Centroid and Space Vectors}    
    \label{fig:centroid-euclidean}
\end{figure}

To set the ground for the experiments, the sample of articles and the category articles are all encoded into the vector space. The centroid vector of the category is calculated by averaging its vectors. We calculate then the euclidean distances between this centroid vector and every other vector in this space. Figure~\ref{fig:centroid-euclidean} shows the results of these calculations. We observe that the distances of the sample articles fall mostly in the interval of distances: $[3.5, 10.5]$ while the category articles have distances to their centroid not exceeding $4$, showing an overlap in $[3.5, 4]$. The upcoming experiments shed more light on how to retrieve the closest articles to the centroid (other than the category vectors), and we highlight how these retrieved articles fall distance-wise.

\subsection{Convex Hull: Geometric Boundaries}
\begin{figure}
    \centering
    \includegraphics[width=.97\linewidth]{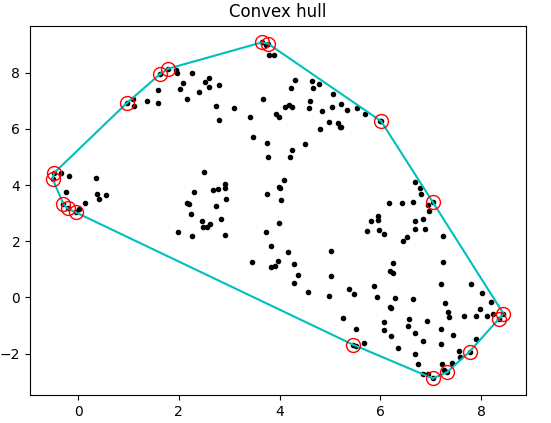}
    \caption{Convex Hull of the Category}    
    \label{fig:convex-hull}
\end{figure}

\begin{figure}
    \centering
    \includegraphics[width=.97\linewidth]{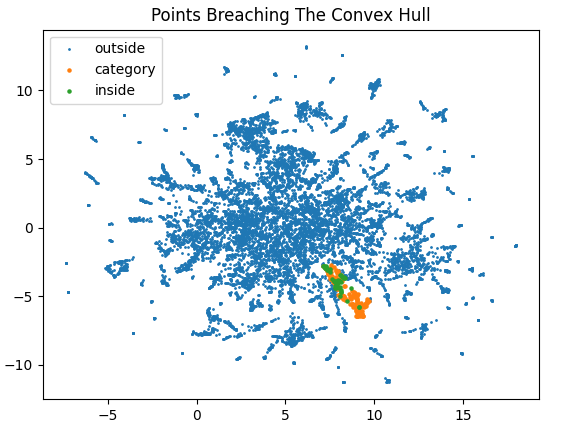}
    \caption{Map of Articles Breaching the Convex Hull}    
    \label{fig:ch-breaching}
\end{figure}

\begin{figure}
    \centering
    \includegraphics[width=.97\linewidth]{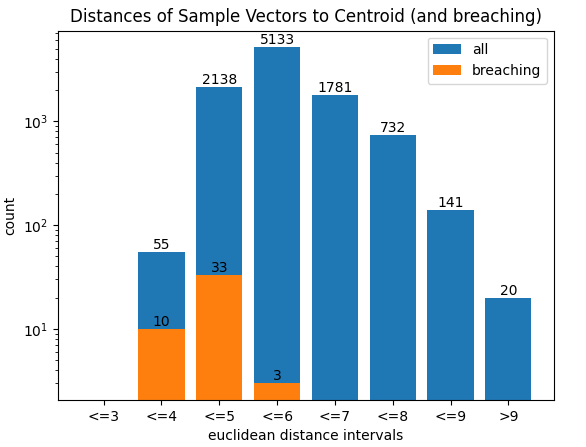}
    \caption{Histogram of Distances of Convex-Hull-Breaching Articles to Category Centroid}    
    \label{fig:ch-breaching-distances}
\end{figure}

In this experiment we construct the convex hull with the encodings of the category (Figure~\ref{fig:convex-hull}). The convex hull is placed on the map of the 10,000 articles and the articles that breach it are recorded (Figure~\ref{fig:ch-breaching}). The distance-to-centroid of each article breaching the convex hull is recorded and plotted in Figure~\ref{fig:ch-breaching-distances}. 

The convex hull successfully identified 10 out of the 55 articles with distances smaller than 4 and 33 with distances smaller than 5.  This outcome showcases a downside to consider, as the convex hull also selected articles positioned farther away from the centroid than the articles it ignored. This is a form of ``blindness” in the method, warranting further exploration and refinement.

\subsection{The Contextual Category in an HNS-World of Articles}

An HNSW is built on the setup vectors (sample articles, category articles, and the centroid vector). We retrieve iteratively all of the category vectors from the position of the centroid vector in the HNSW. In this process, we consider the category vectors as a fishnet to retrieve within any other non-category vector that is closer to the centroid vector than the farthest category-vector. 

\begin{figure}
    \centering
    \includegraphics[width=.97\linewidth]{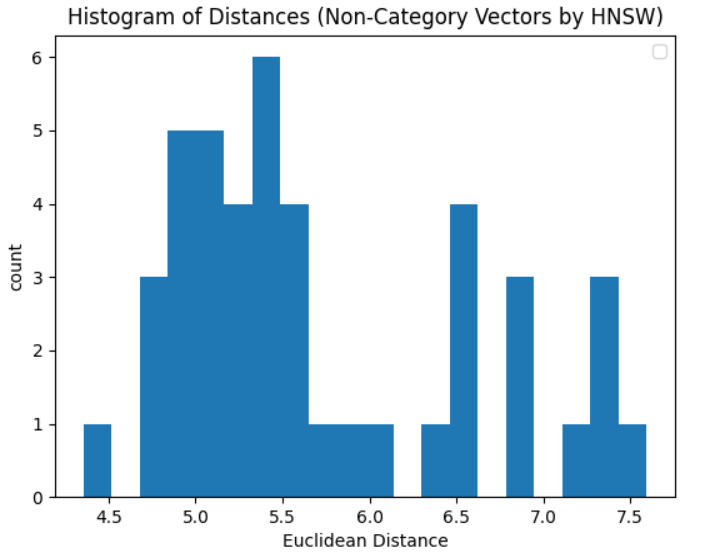}
    \caption{Histogram of Distances of HNSW-Retrieved Articles (non-category) to Category Centroid}    
    \label{fig:hnsw-hist}
\end{figure}

We calculated the distances of the retrieved non-category vectors to the category centroid (Figure~\ref{fig:hnsw-hist}). We observe that the built HNSW jumped over articles closer to the centroid (only one with $d<4.5$) and retrieved farther ones (six with $d=5.4$). This indicates that HNSW follows a pattern similar to the convex hull technique. This similarity arises from the implicit mapping performed by HNSW, resulting in a mapping structure mirroring the loss of information already seen in the convex hull technique.

\subsection{High-Dimensional Latent-Space Filter}
As described in \ref{method:filter}, the 75th percentile distance from the centroid vector to articles of the category $d_{c}$ is 3.151 (Figure~\ref{fig:category-to-centroid}). The median of the distances in the non-category vectors $median(set(d_{ea}))$ is 5.447. By numerical application, Equations 1 and 2 are as follows:

\begin{figure}
    \centering
    \includegraphics[width=.97\linewidth]{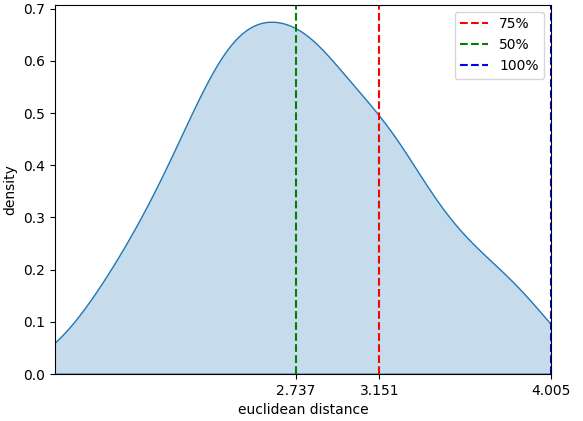}
    \caption{Distances of Category Articles to Category Centroid}    
    \label{fig:category-to-centroid}
\end{figure}

\begin{equation*}
    k = \frac{-ln(0.5)}{median(set(d_{ea}))-d_{c}} = 0.302
    \label{eq:k-num}
\end{equation*}

\begin{equation*}
    RP(d_{c}=3.151, d_{ea}) = 100*e^{-0.302*(d_{ea}-3.151)} 
    \label{eq:rp-num}
\end{equation*}

\begin{figure}
    \centering
    \includegraphics[width=.97\linewidth]{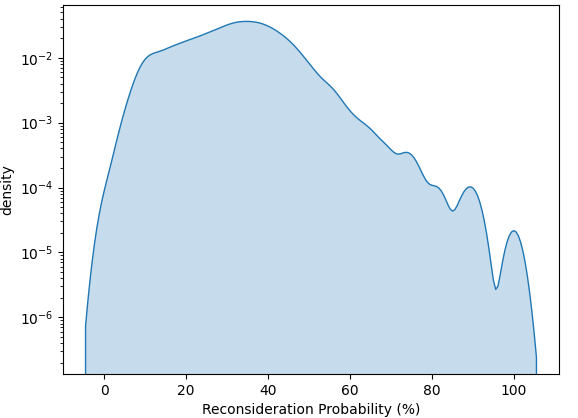}
    \caption{RP of Sample Articles to Category Centroid}    
    \label{fig:sample-to-centroid}
\end{figure}

We calculate the RP values of every articles in the sample (Figure~\ref{fig:sample-to-centroid}). The distances of these articles $d_{ea}$ are in the range of $[3.143, 11.737]$:
\begin{itemize}
    \item For the minimum value: $RP(3.151, 3.143)=100.241\%$ (saturated to 100\%)
    \item For the median value: $RP(3.151, 5.447)=49.988\%$
    \item For the maximum value: $RP(3.151, 11.737)=7.479\%$
\end{itemize}

\begin{figure}
    \centering
    \includegraphics[width=.97\linewidth]{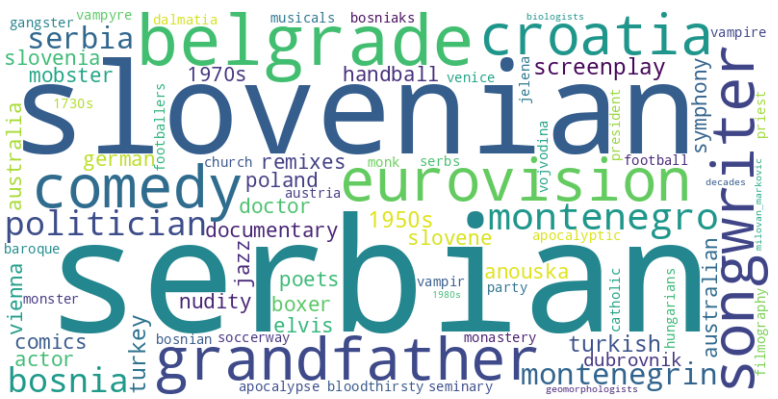}
    \caption{Wordcloud of Articles with $RP>75\%$}    
    \label{fig:rp75-keywords}
\end{figure}

We take the articles with $RP>75\%$, extract their keywords and plot their wordcloud (Figure~\ref{fig:rp75-keywords}). The filter retrieved 20 articles with this threshold. Table~\ref{tab:rp-75} shows the breakdown of closest 5 retrieved items. The category ``Serbian Films" is similar in context to art topics from the Balkans. This explains the presence of articles about writings from the ex-Yugoslavian countries. To test the variations in centroid vector, we sample the data into 100 samples. The mean distance between the centroid vector and the 80\% sample centroid vector is 0.102 with a standard variation of 0.026. This shows an initial stability in the category.

\begin{table}
    \renewcommand{\arraystretch}{1.5}
    \centering
    \caption{Closest 5 Filter-Retrieved Articles (Category: \textit{Serbian Films})}
    \label{tab:rp-75}
    \begin{tabular}{|l|l|l|} 
    \hline
    \textbf{RP (\%)} & \textbf{Title} & \textbf{Keywords}\\
    \hline
    100.00 & Croatia in Eurovision 2006 & eurovision serbia montenegro\\
    \hline
    91.216 & Đani Pervan & bosnian musician songwriter\\
    \hline
    90.993 & Vranjic & venice church\\
    \hline
    90.499 & Vinci Vogue Anžlovar & vampyre blog slovenian\\
    \hline
    89.713 & Jovan Cvijić & serbian ethnological historic\\
    \hline 
    \end{tabular}
\end{table}


\subsection{Bonus: Hierarchical Vectors Effect on Cluster Cohesion}
\begin{figure}
    \centering
    \includegraphics[width=.97\linewidth]{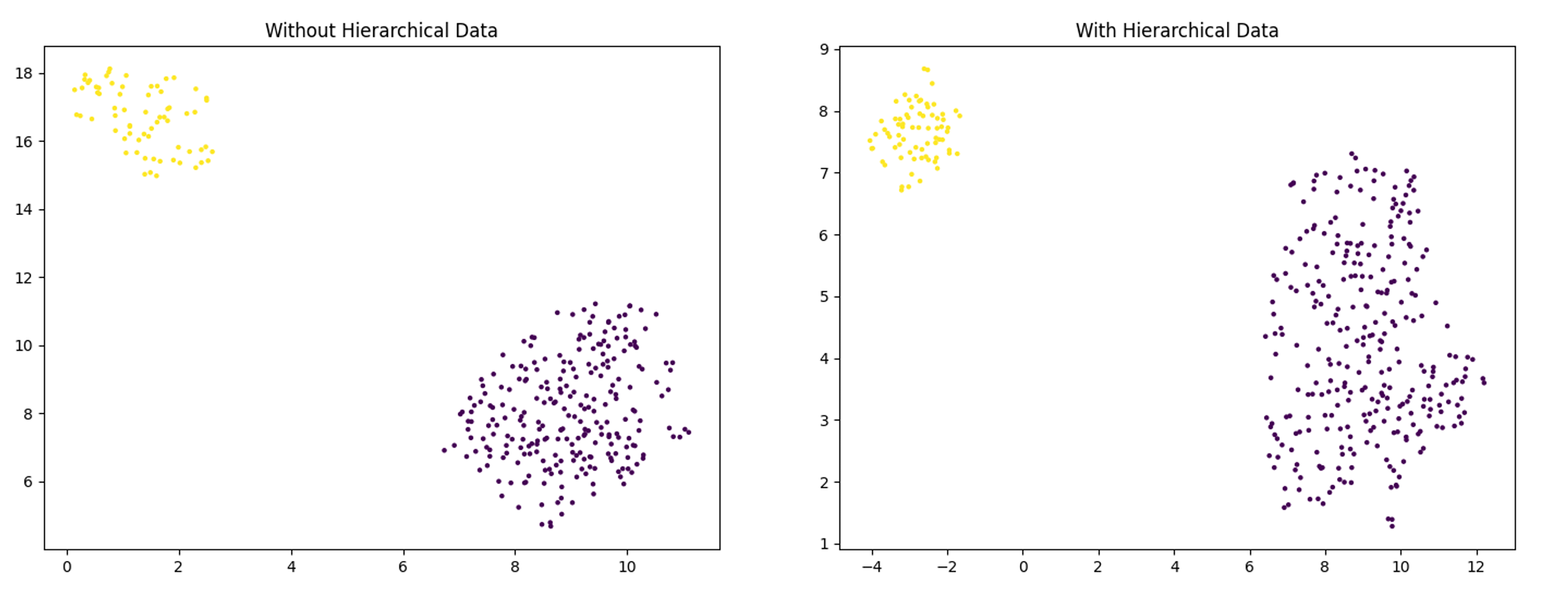}
    \caption{ 2D Mappings of two different categories (left: without hierarchical vectors, right: with them).}    
    \label{fig:hier-vec}
\end{figure}

The presence of hierarchical structures in the form of categories and sub-categories raises the question regarding their utility. One hypothesis worth exploring is whether the incorporation of hierarchical vectors into clusters can reinforce their internal coherence, rendering them more distinct from one another. To quantify this distinction, the Silhouette score~\cite{silhouette} serves as a reliable metric. To initiate the clustering process, two distinct clusters were created based on articles encoded with BERT, resulting in an initial Silhouette score of 0.23. Subsequently, hierarchical vectors were calculated to represent sub-categories (Figure~\ref*{fig:hier-vec}). Each sub-category’s vector was computed as the average vector of its constituent encoded articles (further elucidated in the subsequent section). These hierarchical vectors were integrated into the clustering process, and the clustering was re-executed. Notably, the new Silhouette score was improved by 13\% (0.26). This shows the impact of incorporating hierarchical vectors, leading to denser clusters.

\section{Conclusion and Future Work}\label{conclusion}
In conclusion, this paper has addressed the goal of improving the efficiency of contextual categorization within hierarchical structures. This work employed the Wikipedia dumps and its categories, along with BERT models as the latent-representation technique. The exploration has focused on the integration of hierarchical vectors and advanced clustering techniques.

The experiments showed the practicality of calculating centroid vectors for category trees. The centroid vector, obtained by averaging all the vectors within a category tree, served as a key item for assessing the proximity of articles to the tree. The relationship between an article’s vector and the centroid vector provided a quantitative basis for evaluating the relevance of an article to a specific category tree. This approach allowed for a reconsideration of category labels based on the calculated distances. Alternative techniques such as convex hulls and HNSWs, although explored, exhibited distortions in the results due to the inherent mapping processes involved. To overcome the loss of information and take advantage of the high-dimensionality of the embeddings, we modulated a filter using the exponential decay function that indicates the Reconsideration Probability. The transformer model processes articles individually, ensuring consistent and undistorted encodings regardless of the number of articles processed and sample size. Therefore, the scalability of the exponential decay function (our modulated filter), leveraging transformer embeddings, due to its mathematical nature, offers efficiency gains compared to the scalability challenges typically associated with convex hull and HNSW algorithms. Finally, utilizing hierarchical vectors for subcategories proved valuable, enhancing the representation of the category tree in the latent space. This was evident in the increased Silhouette score, indicating a clearer categorization structure. This utilization is not limited to our case (Wikipedia data), but extends to any form of dataset hosting such hierarchies.

Future research should refine hierarchical vector integration, develop specialized clustering algorithms for complex structures, scale experiments to larger datasets, explore new content categorization tools, assess their impact on platforms like Wikipedia and other databases more scientifically oriented, and enhance categorization systems' precision and utility from a contextual perspective.

\section*{Acknowledgement}
This paper has received funding from the state of Bavaria in the context of project SEMIARID, funding no. DIK-2104-0067//DIK0299/01

\bibliographystyle{IEEEtran}
\bibliography{references}

\end{document}